\documentclass{article}

\usepackage[preprint]{nips_2018}

\usepackage{times}
\usepackage{latexsym}

\usepackage{url}
\usepackage{natbib}
\usepackage{graphicx}
\usepackage{booktabs}
\usepackage{amsfonts}
\usepackage{amsmath}

\newcommand{\quotes}[1]{``#1''}
\newcommand{\kbent}[1]{\textsc{#1}}
\newcommand{\kbrel}[1]{\textit{#1}}

\newcommand{\vect}[1]{\boldsymbol{#1}}

\newcommand{\dsvie}{KBV_{IE}}
\newcommand{\dsvkb}{KBV}

\title{Populating Web Scale Knowledge Graphs using Distantly Supervised Relation Extraction and Validation}

\author{
Sarthak Dash\thanks{Equally contributed} , Michael R. Glass\footnotemark[1] , Alfio Gliozzo, Mustafa Canim \\ 
IBM Research AI \\ 
\texttt{\{sdash, mrglass, gliozzo, mustafa\}@us.ibm.com} \\ 
}

\begin{document}

\maketitle

\begin{abstract}
  In this paper, we propose a fully automated system to extend knowledge graphs using external information from web-scale corpora. The designed system leverages a deep learning based technology for relation extraction that can be trained by a distantly supervised approach. In addition to that, the system uses a deep learning approach for knowledge base completion by utilizing the global structure information of the  induced KG to further refine the confidence of the newly discovered relations. The designed system does not require any effort for adaptation to new languages and domains as it does not use any hand-labeled data, NLP analytics and inference rules. Our experiments, performed on a popular academic benchmark demonstrate that the suggested system boosts the performance of relation extraction by a wide margin, reporting error reductions of 50\%, resulting in relative improvement of up to 100\%. Also, a web-scale experiment conducted to extend DBPedia with knowledge from Common Crawl shows that our system is not only scalable but also does not require any adaptation cost, while yielding substantial accuracy gain.
\end{abstract}

\setlength{\abovedisplayskip}{1.1pt}
\setlength{\belowdisplayskip}{1.1pt}
\addtolength{\parskip}{-0.5mm}

\pretolerance=5000
\tolerance=9000
\righthyphenmin=4
\lefthyphenmin=4

\hyphenation{KBC}
\hyphenation{KBV}

\section{Introduction}


Knowledge graphs (KGs) are widely used in question answering and dialogue systems. Minimizing the error rate in these graphs without sacrificing coverage of entities and relationships is essential for improving the quality of these systems. In this paper we focus on the problem of identifying relations among entities found in a large corpus with the goal of populating a pre-existing KG. 
\emph{Relation Extraction (RE)} from text is described as inducing new relationships between pre-identified entities belonging to a predefined schema. Expanding the size and coverage of a knowledge graph with relation extraction is a challenging process as it introduces noise and oftentimes requires a manual process to clean it.

Human experts are able to perform very well on the task of recognizing relations in text because they make use of background knowledge and inference to validate them. Yet, computers are not as proficient as humans when it comes to performing this task. 





\begin{figure}[ht]
   \centering
   \includegraphics[width=.5\linewidth]{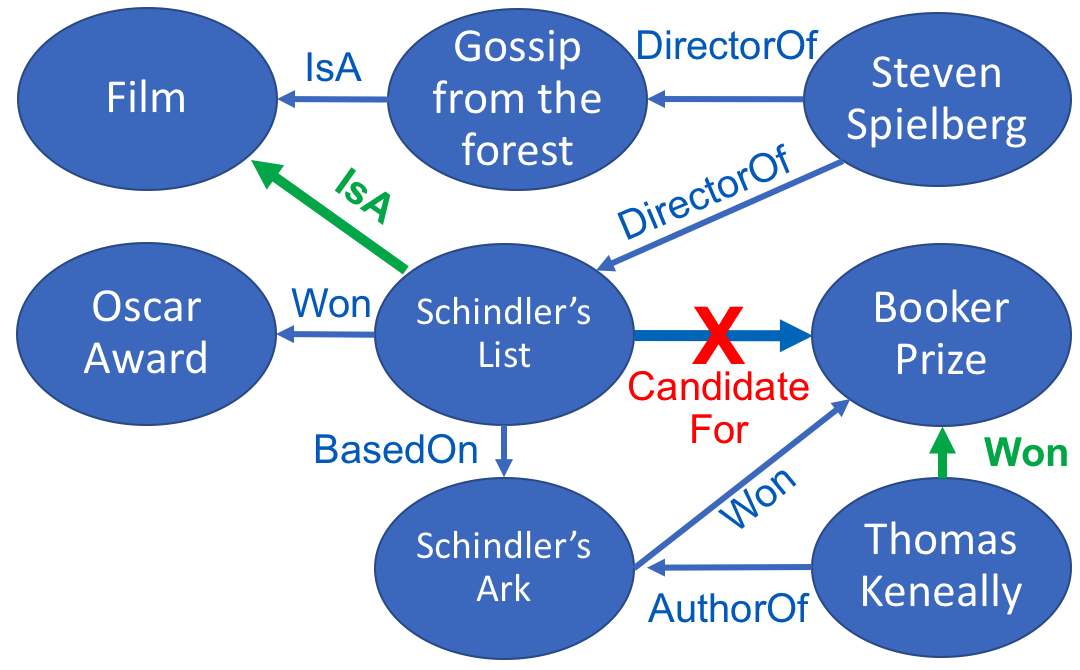}
   \caption{Link Prediction and Knowledge Base Validation example.}
   \label{fig.kbc_example2}
\end{figure}

For example, an automatic system might have reasonably high confidence in the relationship \kbent{``Schindler's List} - \kbent{candidateFor} - \kbent{Booker Prize''}  from the text 
\textit{``Thomas Keneally has been shortlisted for Booker Prize in four different occasions, in 1972 for The Chant of Jimmie Blacksmith, Gossip from the Forest in 1975, and Confederates in 1979, before winning the prize in 1982 with Schindler's Ark, later turned into the Oscar Award winning film Schindler's List directed by Steven Spielberg.''}
However, as illustrated in Figure \ref{fig.kbc_example2}, other extracted relationships might contradict this, such as the fact that because  \kbent{Steven Spielberg} directed \kbent{Schindler's list}, it follows that \kbent{Schindler's list isA Film} and therefore it cannot be  \kbent{candidateFor} the  \kbent{Booker Prize}, which is a literary award. 
The first type of inference is equivalent to identify a new relation in a KG, and it is typically referred to as \emph{link prediction}, as illustrated by Figure \ref{fig.kbc_example2}. The second inference step is equivalent to assess the confidence of an existing relation in the KG and it is typically referred to as Knowledge Base Validation (KBV). Both processes are very intimately related and interfere with each other. In the example before, we needed to infer that  \kbent{Schindler's List isA Film} from the explicit information in order to detect the fact that \kbent{Schindler's List} cannot be candidate for the \kbent{Booker prize}. 

Humans are able to reconcile inconsistencies like these at an almost subconscious level, resulting in improved perception capabilities. Unfortunately, this is not the case for most AI systems and this is one of the main reasons why pure NLP based approaches, whether pattern based or deep learning based, typically perform poorly on this task.

In this paper, we present an approach that overcomes the aforementioned problem while offering a scalable solution to extend large Knowledge Graphs from web-scale corpora. It consists of two main components:
\textbf{Relation extraction,} a deep learning based distantly supervised system to detect relations from text;
\textbf{Relation validation,} a deep learning based knowledge base validation component able to spot inconsistencies in the acquired graphs and improve the global quality.
In order to operate these components, the only required input is a partially-populated KG and a large scale document corpus. In our experiments, we used DBpedia and Freebase for the KG and Common Crawl web text and New York Times news articles for the document corpora. 

To implement the \emph{RE component} we applied a state of the art distantly supervised relation extraction system, that is capable of recognizing relations among pre-identified entities using a deep neural network approach \citep{glass2018implicit}. \emph{Entity recognition} is simply achieved by using a dictionary matching approach in a large corpus without requiring an \emph{entity detection and linking} system. As for the \emph{Relation Validation (RV) component}, we used a deep neural network approach trained from the same KG as well as from the relations identified from text, adopting Knowledge Base Completion (KBC) strategies.

The main contribution of this paper is that we show how combining distantly supervised solutions for RE with KBC techniques trained on top of their output can largely boost the overall RE accuracy, providing a scalable yet effective solution to extend their coverage. We describe a system combining those two approaches in a single framework, and we apply it to the problem of extending KG from web-scale corpora. In previous art, KBC has been applied to hand-crafted knowledge bases and not to the result of the information extraction system. We empirically show how this combination improves the quality of the induced knowledge by a large margin, improving the state of the art in a scalable manner.

We tested our approach on three different KBP benchmarks: extending Freebase with knowledge coming from NYT, extending DBpedia with knowledge coming from Common Crawl and refining the result of pattern based Information Extraction systems used for the Never Ending Language Learning (NELL) task. 
Our experiments show that the \emph{Validation} step boosts the performance of RE by a wide margin, reporting error reductions of 50\%, sometimes resulting in relative improvement of up to 100\%. 

The rest of the paper is structured as follows. The related work section describes the background in the area of RE and KBC, as well as alternative approaches such as the application of probabilistic logic to the validation of KBs. We then introduce our approach and provide a description of the RE system we use for our experiments. The evaluation section describes the benchmarks and provides an extensive evaluation of our framework, followed by an analysis of the reasoning behind its effectiveness. Finally, we summarize the main research result and highlight possible directions for future work.

\begin{figure}[ht]
   \centering
      \includegraphics[width=0.6\linewidth]{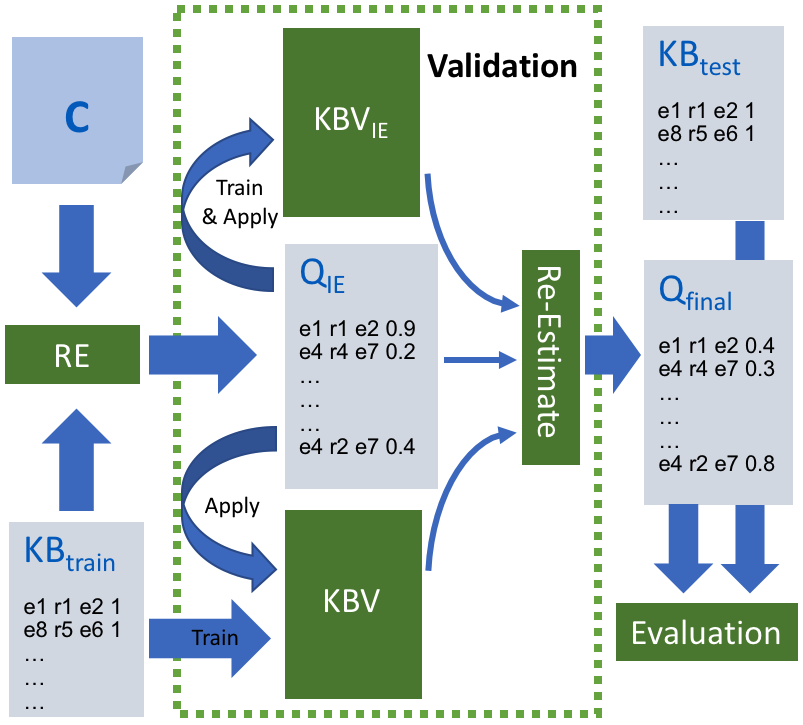}
   \caption{Pipeline for our RE solution.}
   \label{DSKV}
\end{figure}

\section{Related Work}\label{relatedWork}




Deep learning has been widely explored for the task of information extraction. Both CNN-based \citep{zeng2014relation} and LSTM-based \citep{xu2015classifying} models have been trained successfully for RE. Recently, cross sentence approaches have been explored by building paths connecting the two identified arguments through related entities \citep{zeng2016incorporating}. The context aggregation approaches of state-of-the-art neural models, max-pooling \citep{zeng2015distant} and attention \citep{lin2016neural} allow multiple contexts to contribute to a predicted relation between two entities. 


The efforts described above to aggregate information from different sentences are clearly a step toward our goal of providing a global assessment of the validity of the recognized relation. However, all the systems above lack the ability to handle global knowledge, for example, derived from sentences involving other related entities, severely limiting their accuracy. One attempt to leverage background knowledge to improve RE for Knowledge Base Population is the Universal Schema of \citet{riedel2013relation}, where a matrix factorization approach uses evidence from both the ontology and text to identify new relations. Universal Schema, by closely integrating the textual and knowledge base evidence, limits the approaches to each. In contrast, by defining a symbolic layer to separate the IE and KBC components, our approach is able to easily accommodate different implementations of either the IE component or KBC component.


Probabilistic reasoning has been explored to validate the output of RE systems, including Markov Logic Networks (MLN) \citep{markovLogic} and Probabilistic Soft Logics (PSL). For example, in the Never Ending Language Learning (NELL) project \citep{carlson2010toward}, PSL attempts to reconcile the output of IE systems, which provide heterogeneous and often contradicting sources of evidence for some relations, with the constraints of the KB \citep{pujara2013knowledge}. However, probabilistic reasoning based approaches require logical statements describing the target knowledge schema such as domain and range constraints or taxonomies and ground truth of manually validated facts, as entity-relation-entity triples, for training. After training is performed, a PSL or MLN system is able to validate statements in a knowledge base, such as detecting inconsistencies. But on large datasets, the systems often suffer from scalability problems.

Fact checking is another line of research related to knowledge base validation. A typical fact checking system gathers more textual evidence for a given proposition through information retrieval, often a web search \citep{gerber2015defacto}. In contrast, our system builds a global model for the entities and relations considering the interactions of the extractions rather than gathering more documents.










On the other hand, KBC technology has been developed to perform a similar function and has been applied to knowledge bases curated by humans.
State of the art KBC approaches are usually deep learning based. They are trained using triples in the input KB as positive examples and generate negative examples by random corruption of the training data. Popular KBC approaches are TransE \citet{bordes2013translating},
RESCAL \citet{nickel2011three},  Neural Tensor Network \citep{socher2013reasoning} and HolE \citep{nickel2016holographic}; whereas newer ones include ConvE \citep{dettmers2017convolutional}, ConvKB \citep{nguyen2017novel}, KBGaN \citep{cai2017kbgan} as well as many others. In this paper we exploit a variant of ProjE \citep{shi2017proje} able to take noisy data with an associated confidence score as an input. This is $\dsvie$, a core component of our KBP system.

\section{Distantly Supervised Relation Extraction and Validation}\label{DSV}

In this section, we describe the architecture of our solution for Knowledge Base Population (KBP). KBP is the task of identifying entities and relations from a corpus, according to a predefined schema. It is illustrated by Figure \ref{DSKV}, representing the architecture of our final KBP solution. It is composed by a distantly supervised Information Extraction system that takes a pre-existing KB and a corpus as an input and generates a list of quads representing induced relations with their associated confidence scores. Its output is then merged with the triples in the pre-existing KG and fed into a KBC deep net to train a KBV system whose goal is to re-assess the generated assertions, providing new confidence scores for each of them. Finally, those scores are aggregated by a final logistic regression layer that provides the final confidence score for each triple. For all those steps, the same KB is always used for training.

More formally, the information extraction component of KBP generates a set of quads (triples with confidence) $Q_{IE} = q_1, q_2, ... , q_n'$ from a corpus of text $C = c_1, c_2, ... , c_m$.
 in the form of sequences of words $c= w_1,  ... , e_1, w_a, ...., w_b, e_2, ... , w_z$ containing two entity mentions $e_1$ and $e_2$ each. 
Quads have the form $q = \langle e_1, r, e_2, s \rangle$ where $e_i \in \mathcal{E}$ are entities found in the corpus, $r \in R$ is a finite set of relations and $s\in[0,1]$ is a confidence score. We define the function $\tau( \langle e_1, r, e_2, s \rangle ) = \langle e_1, r, e_2 \rangle$ to ignore the confidence of a quad, forming a triple.
Since $KB$ is typically the Abox of an hand crafted ontologies, we assume all the confidence scores of quads in $KB$ being equal to 1.
For each context $c \in C$ the Entity Detection and Linking (EDL) function $\psi(c) = <e_1 , e_2>$ returns the two entities contained in it. In our current implementation EDL is implemented by a simple string match w.r.t. the entities in the KB, however, it could be also replaced with more advanced EDL solutions if available. 
For each entity $e\in V$, the function $\psi(e)$ returns all possible contexts where the entity $e$ appears in the corpus and $\psi(e_1, e_2)$ returns all contexts containing both. 
The RE process consists on applying a deep net to the context returned by $\psi(e_1, e_2)$ for every pair of entities that co-occur in the corpus. The result of the application of RE to a context is a list of quads  $q = \langle e_1, r_i , e_2, s_i \rangle for all r_i \in R$, where $s_i$ represents the confidence of the system on the detection of the relation $r_i$ in one or more contexts in $\psi(e_1, e_2)$, where the two entities co-occur in the corpus. Obviously, most of the relations will have very low scores since all the relations are explored and returned for each pair. 




The RE step takes into account mostly information coming from the corpus for each entity pair to predict the relations, if any, between them. It doesn't take into account global information provided by the structure of the KG. The Relation Validation component is designed to overcome this problem. 
It is formally described as a function $KBV: \mathcal{E} \times R \times \mathcal{E} \mapsto \mathbb{R}$. For any triple produced by IE, ($\tau(q) : q \in Q_{IE}$), KBV returns a confidence score. 

The KBV system is to be trained from a knowledge graph $KB$ consisting of a set of quads. In this paper, we experimented with two different ways of training, producing two-component systems: (a) $\dsvkb$, using the ground truth from the knowledge graph $KB_{train}$, and (b) $\dsvie$ using the output of information extraction $Q_{IE}$. The result is two different functions returning different confidence scores when applied to the same triple.



The three confidence scores generated from IE and by applying $\dsvkb$ and $\dsvie$ to every triple from $Q_{IE}$ are then aggregated using a confidence re-estimation layer trained on a validation set to provide a final confidence score, generating the final output $Q_{final}$.

In the following subsection we will describe the Distantly Supervised RE approach and the Knowledge Base Validation step into details.


\subsection{Relation Extraction}\label{sec.deeplearningarch}

We use knowledge-level supervision, sometimes called distant supervision, to generate training needed for deep learning based RE systems from a KG and an unannotated corpus.  

To this aim, we first match all entities in $KB_{train}$ to gather their context sets. That context set provides all the sentences that contain two entity mentions. If those two entities are related by some relation in the input KG, they become positive examples for that binary relation. 

We then use all the context sets collected from the corpus to train a deep learning based RE classifier. We use the system of \citet{glass2018implicit} based on the PCNN model from NRE \cite{lin2016neural}.

It is worthwhile to notice here that for each entity pair we predict a probability distribution for all the possible relations in our KB. To avoid generating a very large list of quads, a confidence threshold is chosen below which quads are discarded before passing to the KBV system. 

After the system is trained, it is applied to all context sets for every pair of entities in the corpus $C$ and generates a set of quads $Q_{IE}$, where for each pair of entities $e_1$ and $e_2$ up to $|R|$ triples are generated and associated with their confidence score. Minimum confidence is set for extracted quads to control the size and quality of the output.

\subsection{Relation Validation}





We implement $\dsvie$ using a deep network inspired by a state of the art KBC approach where we modified the loss function in order to take into account the fuzzy truth values provided by the output of IE. 
This network considers a set of quads $Q_{IE}$ as the probabilistic knowledge graph for training and learns a function $\dsvie(\langle e_1, r, e_2 \rangle)$ 
that returns a confidence score $s$ for the triple at hand. This score is informed by the global analysis of the knowledge graph $Q$ differently from the $RE$ that uses the evidence from the corpus $\psi(e_1, e_2)$ for the same purpose.







KBC algorithms are trained from a set of triples $T$, usually produced manually, wherein each entry $t \in T$ comprises two entities $e_1, e_2$ and a relation $r$. The KBC system assigns tensors to the entities and relations and trains them by exploiting a Local Closed World Assumption.




\begin{figure}[ht]
   \centering
   \includegraphics[width=0.7\linewidth]{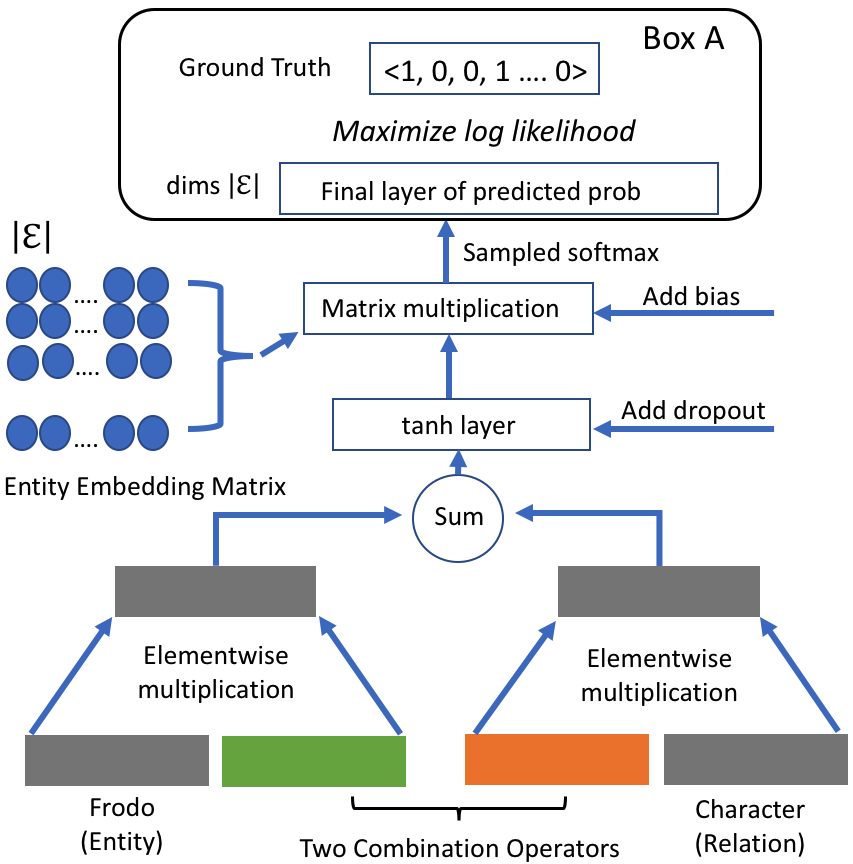}
   \setlength{\belowcaptionskip}{-10pt}
   \caption{Base ProjE softmax architecture for KBC.}
  
   \label{fig.proje_base}
\end{figure}
 

In this work, we use a state of the art model for KBC, called ProjE softmax \citep{shi2017proje}.  A block diagram architecture of such a model is shown in Figure \ref{fig.proje_base}. 
The network is trained for each triple $t$ in the training data by providing an input vector representation for the subject and the relation, while the output of the network exploits a one-hot representation encoding the probability for each possible object in $\mathcal{E}$. Negative examples are provided by a random sampling of the objects.
 
However, this approach cannot be directly applied to implement $\dsvie$ because many triples extracted by IE are actually not true. This is usually reflected by a lower confidence score associated with the triple.
To overcome this issue, we modified the loss function described in Figure \ref{fig.proje_base} (Box A) to use confidence scores, rather than labels, following an approach proposed for Computer Vision in \citet{gong2013deep}.\vspace{0pt} 
Let us assume that the inputs are $e_1$ and $r$, and the system needs to predict appropriate $e_2$. 
Let $\vect{v}^{e_1,r}$ (of dimensions $|\mathcal{E}|$ - number of entities in vocabulary) represent the final layer of predicted probabilities corresponding to input entity $e_1$ and input relation $r$. Define a vector $\vect{s}^{e_1,r}$ of dimensions $|\mathcal{E}|$ that uses the input confidence scores as follows,
\begin{equation}
    \vect{s}^{e_1,r}_i = 
    \begin{cases}
    s, & \text{if}\ \langle e_1,r,e_i,s \rangle \in Q \\
    0, & \text{otherwise}
    \end{cases}
\end{equation}
Recall that $s$ represents the confidence score for the quad $\langle e_1,r,e_i,s \rangle \\ \in Q$. The modified loss function is now the cross-entropy between the confidence vector and the prediction vector. 
\begin{equation} \label{kbv_loss}
    \mathcal{L} = -\frac{1}{|Q|}\sum_{q \in Q} \sum_{i=1}^{|\mathcal{E}|} \vect{s}^{e_1,r}_i \log \vect{v}^{e_1,r}_i    
\end{equation}

In Equation (\ref{kbv_loss}), the $\vect{s}$ vector is now a vector of confidence scores (rather than an one-hot encoding). 

After the network is trained, it can be used for both link prediction (i.e. generating the object from a subject and relation input) or validation (i.e. assessing the validity of a new triple composed of known entities and relations). In this paper, we explore the second option.

The predictions of $\dsvkb$ and $\dsvie$ make use of the embeddings of entities that are determined by the training set. Embeddings for an entity can be effectively trained only when the number of triples in which the entity appears meets some minimum threshold: three in our work.  The KBC system cannot provide a confidence estimate for triples involving entities that do not occur in the training set or occur more rarely than the minimum threshold. This is a critical limitation of typical KBC systems, which can only predict new relations between existing entities in the knowledge base. $\dsvie$ removes this limitation by using the output of the IE system for training, which can include new entities.




\subsection{Confidence Re-Estimation}
\vspace{0.3cm}

The confidence scores $s_{\xi}$ from the three systems $\xi \in \{IE, \dsvie, \dsvkb\}$ are combined to produce a final confidence for each triple $\tau(q) : q \in Q_{IE}$, yielding $Q_{final}$. This step uses a simple logistic regression, typically trained on a validation set separate from the training set, but it can also use the training set itself.

We use four groups of features based on the confidence of each system: the raw confidence itself $f^{raw}_{\xi} = s_{\xi}, \xi \in \{IE, \dsvie, \dsvkb\}$, the logit of the confidence $f^{logit}_{\xi} = \log(\frac{1}{s_{\xi}}-1)$, and binary features for what range the confidence is in $f^{bin}_{\beta}, \beta \in \{[0,0.2), [0.2,0.4),$ 
$[0.4,0.6), [0.6,0.8), [0.8,1.0]\}$. If one of the entities occurs too few times, either in 
$T_{KB}$ for $\dsvkb$ or $Q_{IE}$ for $\dsvie$ it will not have an embedding and therefore will not have a score from $KBC$. In this case the re-estimation uses a binary feature to indicate the confidence from the system is missing $f^{missing}_{\xi}, \xi \in \{\dsvie, \dsvkb\}$.  



We also introduce a binary feature to indicate the relation in the triple to enable learning a per-relation bias $f^{rel}_{r}, r \in R$.
Finally, we form quadratic features by adding a feature for the product of every pair of features. This captures basic interactions between features. L1 regularization is applied to reduce overfitting. 

\vspace{-0.2cm}
\section{Evaluation}\label{evaluation}

\begin{table}
\footnotesize
\center
\caption{Knowledge Base Population dataset statistics} \label{tab:dataset}
\begin{tabular}{ rrrr } 
 \toprule
  & NYT-FB & CC-DBP & NELL-165 \\ \midrule
  $|Q_{IE}|$  & 23,687 & 6,067,377 & 1,030,600  \\
  $|KB_{KB}|$ & 15,417 & 381,046 & 2,928 \\ 
  $|\mathcal{E}|$ & 17,122 & 545,887 & 820,003 \\ 
  $|R|$ & 13 & 298 & 222 \\ 
  \bottomrule
\end{tabular}
\end{table}

\begin{figure*}[ht]
   \centering
    \begin{minipage}{.33\textwidth}
   \centering
   \includegraphics[width=1\linewidth]{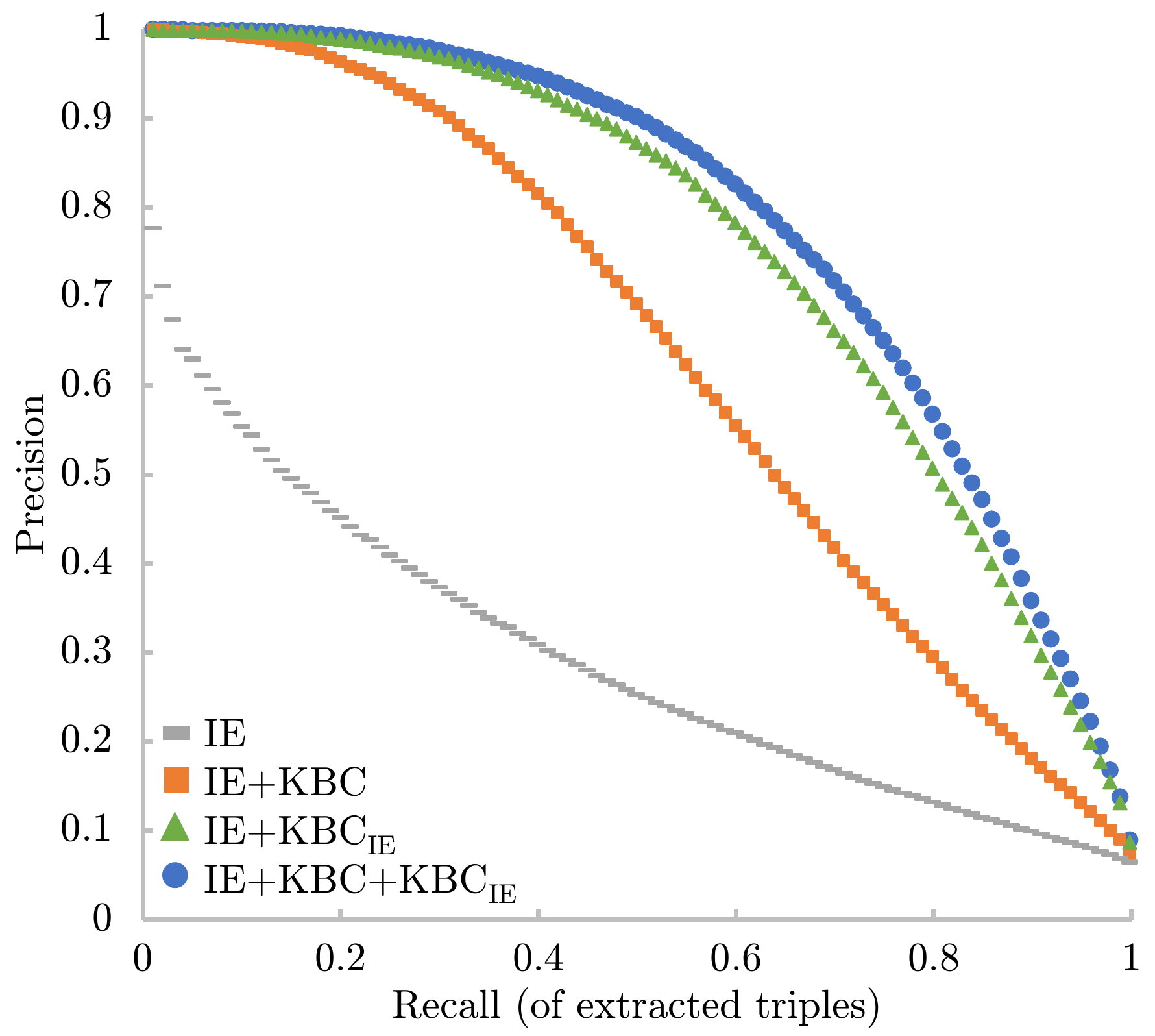}
   \caption{Precision / Recall curves for CC-DBP}
   \label{fig.ccdbpPR}
\end{minipage}%
\begin{minipage}{.33\textwidth}
   \centering
   \includegraphics[width=\linewidth]{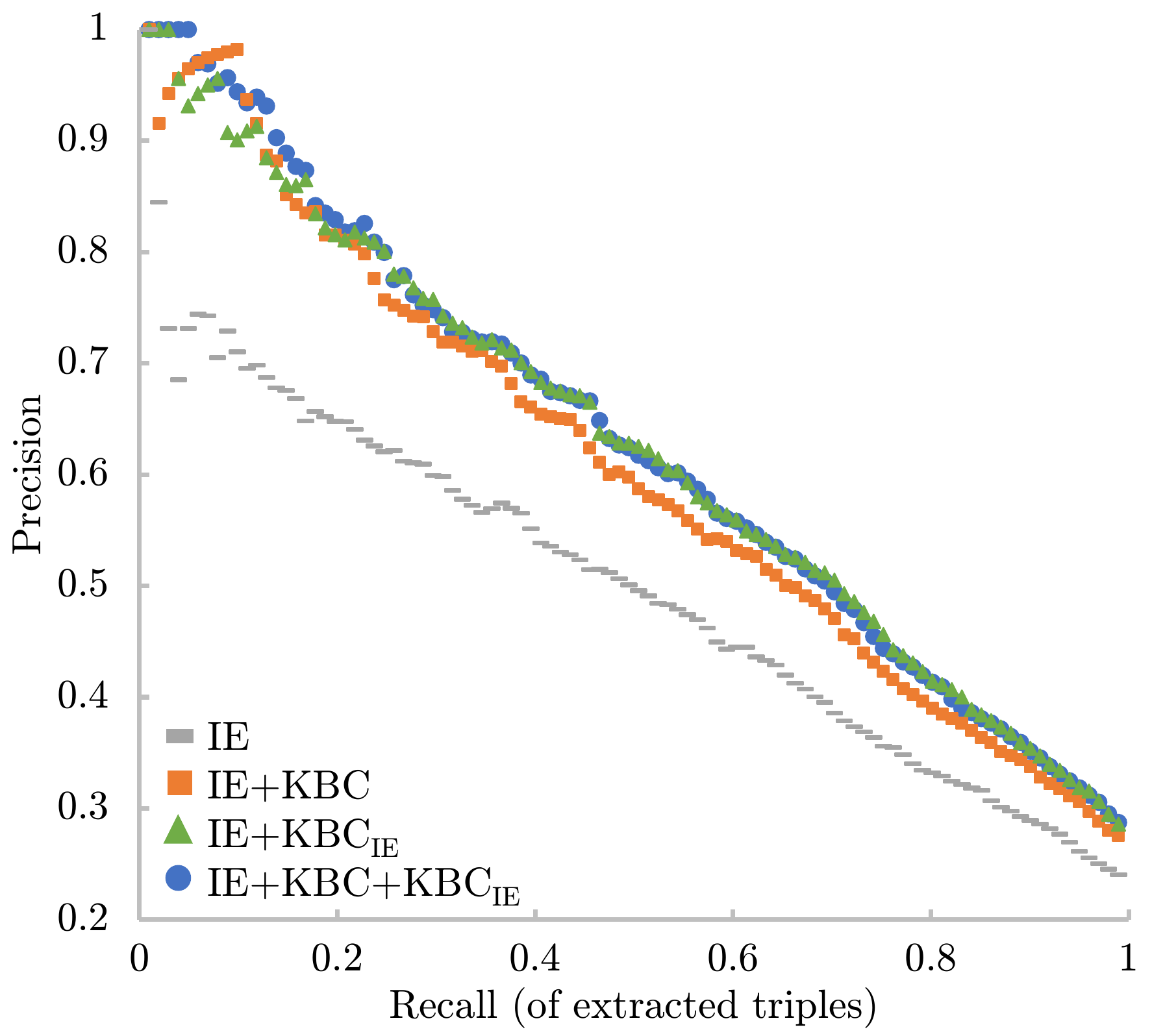}
   \caption{Precision / Recall curves for NYT-FB}
   \label{fig.nytfbPR}
\end{minipage}
 \begin{minipage}{.33\textwidth}
   \centering
   \includegraphics[width=\linewidth]{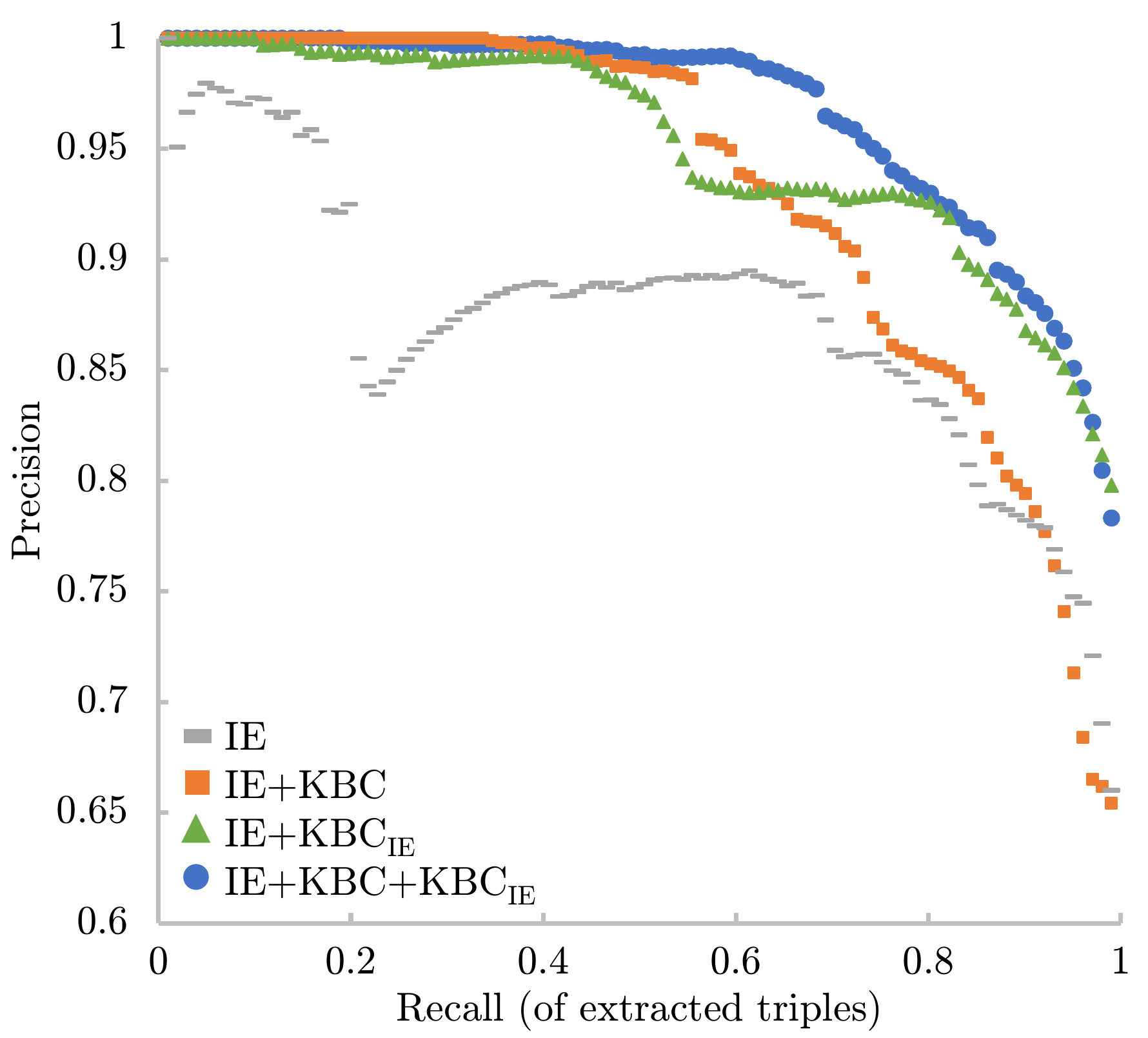}
   \caption{Precision / Recall curves for NELL}
   \label{fig.nellPR}    
\end{minipage}
\end{figure*}


We tested our approach on three different KBP benchmarks: extending Freebase with knowledge coming from NYT, extending DBpedia with knowledge coming from Common Crawl and refining the result of pattern based Information Extraction systems used for the Never Ending Language Learning (NELL) task.  We choose the first task to provide a comparison with the existing state of the art methods for RE, while we use the second benchmark to show the scalability aspect of our approach. We chose the third task to compare the performances of our KBC approach w.r.t. previously made alternative attempts to refine the output of IE system using probabilistic reasoning methods. Benchmarks are described in subsection \ref{benchmarks}, evaluation is reported in Subsection \ref{sec.results} and an analysis of the results is provided in \ref{sec.analysis}.

\subsection{Benchmarks}\label{benchmarks}
For our experiments, we used the following evaluation benchmarks (dataset statistics are summarized by Table \ref{tab:dataset}):

\noindent
\textbf{NYT-FB:} Extending Freebase with New York Times articles is a standard benchmark for distantly supervised RE, developed by \citet{riedel2010modeling} and used in many subsequent works \citep{hoffmann2011multiR,surdeanu2012multi,zeng2015distant}. The text of New York Times was processed with the Stanford NER system and the identified entities linked by name to Freebase. The task is to predict the instances of 56 relations from the sentences mentioning two arguments. The state-of-the-art for this dataset is NRE's (Neural Relation Extraction) PCNN+ATT model (Piecewise Convolutional Neural Network with Attention) \citep{lin2016neural}. 

\noindent
\textbf{CC-DBP:} Extending DBpedia with Web Crawls. This is a web-scale knowledge base population benchmark that was introduced by \citet{ccdbp} and has been made publicly available.
 It combines the text of Common Crawl
with the triples from 298 frequent relations in DBpedia \citep{dbpedia}. 
Mentions of DBpedia entities are located in text by gazetteer matching of the preferred label. This task is similar to NYT-FB but it has a much larger number of relations, triples, and textual contexts.

\noindent
\textbf{NELL:} Never-Ending Language Learning (NELL) \citep{carlson2010toward} is a system that starts from a few \quotes{seed instances} of each type and relation, which then uses to extract candidate instances from a large web corpus, using the current facts in the knowledge base as training examples.  
The NELL research group released a snapshot of its accumulated knowledge at the 165th iteration, hereby referred to as NELL-165 consisting of a set of triples with associated confidence scores coming from different extractors. Later, \citet{knowledgeGraphMLN} provide a manually validated set of triples divided into train and test.

In the case of NELL, the ground truth is in the form of manually validated extractions provided by \citet{knowledgeGraphMLN}. In the cases of CC-DBP and NYT-FB, the ground truth for a triple is determined by its presence or absence in DBpedia or Freebase respectively. This is a positive-unlabeled evaluation and therefore precision is underestimated.
In all cases, the recall is the correct percent of triples that were extracted by the IE system above minimum confidence. 
This recall basis is logical in the case of KBV, but note that $\dsvkb$ or $\dsvie$ could also be used to predict triples outside the set extracted by an IE system.



\subsection{Results}\label{sec.results}


To understand the impact of each component for our Distantly Supervised Relation Extraction and Validation System system ($RE$, $\dsvkb$ and $\dsvie$) we report an ablation analysis: we train the re-estimation component from a subset of the features and plot the precision/recall curve. 

RE performance is illustrated by the three gray lines in Figure \ref{fig.nellPR}, \ref{fig.ccdbpPR} and \ref{fig.nytfbPR}. It is worthwhile to notice that we used our deep learning based approach on CC-DBP and NYT-FB. This system in itself provides state of the art results in those tasks. For NELL the reported results are obtained by using triples provided by the NELL organizers and generated with their system.

Then, we trained both $\dsvkb$ and $\dsvie$ for all the benchmarks. The training set for $\dsvkb$, $KB_{train}$, is derived by the riples validated by humans for NELL, whereas it consists of the intersection of $Q_{IE}$ with  $KB_{train}$  for both CC-DBP and NYT-FB. 
In all cases, $\dsvie$ is trained on the output of the IE systems. Then we apply both systems to validate $Q_{IE}$, the output of the IE system, generating two additional confidence scores.  

Precision / Recall curves for those experiments are given in Figure \ref{fig.nellPR}, \ref{fig.ccdbpPR} and \ref{fig.nytfbPR}. 
Both $\dsvie$ and $\dsvkb$ largely improve the ranking of output triples, promoting the right ones on top.  Remarkably, $\dsvie$ tends to perform better than $\dsvkb$ in spite of the fact that the latter uses manually curated training triples from $KB_{train}$, while the former uses the noisy output of the RE system. 


Finally, we combined all three output scores: $IE$, $\dsvie$ and $\dsvkb$. Results are reported by the blue line in the three PR curves. The blue line is consistently above all the other lines, showing that there is some complementary signal from the three features. However, this improvement is marginal compared to what provided by $\dsvie$ alone. Table \ref{tbl.evaluationSummary} provides the AUC for all the systems described above.

\begin{table}
\footnotesize
\center
\caption{Results: Area Under Precision/Recall Curve (AUC) on KBP Datasets} \label{tbl.evaluationSummary}
\begin{tabular}{ rrrr } 
 \toprule
 Approach & NYT-FB & CC-DBP & NELL \\ \midrule
  $IE$  & 0.499 & 0.294 & 0.872  \\  
  $IE, \dsvkb$ & 0.609 & 0.636 & 0.931 \\  
  $IE, \dsvie$ & 0.629 & 0.760  & 0.951 \\  
  $ALL$ & \textbf{0.630} & \textbf{0.785} & \textbf{0.966} \\ 
  \bottomrule
\end{tabular}
\end{table}

The NYT-FB experiments shows clearly that our approach outperform state of the art solutions for Distantly Supervised RE, represented by the performance of the RE component alone, by a large margin of 0.13 AUC improvement. However, NYT-FB is a relatively small benchmark, and might not be a realistic setup to benchmark large scale solutions for KBP. 

To demonstrate the scalability of our approach, the CC-DBP experiment is performed on a much larger web scale corpus with hundreds of different relations. In these settings, the improvements over state of the art Distantly Supervised RE solutions are even higher, reporting an absolute increase of AUC of 0.491, reflecting on a relative improvement of 167\% . This extraordinary boost in performances can be explained by the fact that larger graphs tend to provide more valuable signal to the KBV process, as demonstrated in the following Subsection.

Finally, the NELL experiment demonstrates how KBV can be an effective alternative to PSL on the task of validating output of IE systems.
It is worthwhile to notice that the best reported result on the task of validating the output triples in NELL is 90.4 AUC, obtained by \citet{pujara2013knowledge} using PSL. This approach requires constraints from the KG schema and a sample of manually validated triples to train from. In our unsupervised settings (i.e. using KBV trained on top of the result of RE only) we achieve an improvement of +0.027 without  even requiring constraints from the ontology. Remarkably, in its supervised settings (i.e. when KBV is also trained from the available manually validated triples) this solution performs much better than the PSL approach, achieving an AUC of 96.6\%. This result is particularly impressive because PSL  requires constraints from the ontology such as taxonomies and domain and range as well as supervised data, whereas KBV does not have any such requirements.

        
\subsection{Analysis}
\label{sec.analysis}

A further analysis considers the improvement in the connectivity of the triples to the other triples in the same group. Our hypothesis is that the KBV will improve the confidence score mostly for statements containing entities that we know many facts about, enabling implicit reasoning.

To test this hypothesis, we define \emph{minimum connectivity} for a triple to be the minimum of the number of triples in which each argument is present. So triples with high minimum connectivity have arguments with KBC embeddings that were influenced by many other triples. We group the triples by their minimum connectivity and calculate the increase in AUC for $IE, \dsvkb, \dsvie$ relative to $IE$ alone for different buckets of triple minimum connectivity. Table \ref{tbl.mincon} shows these results.

\begin{table}
\footnotesize
\center
\caption{KBP's increased AUC by minimum connectivity group} \label{tbl.mincon}
\begin{tabular}{ rrrr } 
 \toprule
 Min. Conn. & NYT-FB & CC-DBP & NELL-\kbrel{Cat} \\ \midrule
       $[1,2)$        & -0.001 & -0.002 & N/A   \\  
    $[2,4)$        & 0.198  & 0.038  & 0.054 \\  
    $[4,8)$        & 0.265  & 0.210  & 0.049 \\  
    $[8,16)$       & 0.442  & 0.460  & 0.036 \\  
    $[16,\infty)$  & 0.377  & 0.634  & 0.084 \\ 
  \bottomrule
\end{tabular}
\end{table}

NYT-FB and CC-DBP, and to a lesser extent NELL, show a consistent picture, with increasing minimum connectivity leading to the largest increases in performance. For NELL we excluded the \kbrel{Cat} relation, which connects an entity to its type since this relation behaves very differently. The NELL \kbrel{Cat} relation increases from 0.925 to 0.997 AUC.


This supports our hypothesis that KBV can improve RE through background knowledge. Since triples with higher minimum connectivity interact with larger amounts of relevant background knowledge.

\section{Conclusion and Future Work}\label{conclusion}


In this paper, we introduced a novel approach to extend the coverage of knowledge graphs, consisting of a combination of Relation Extraction and Knowledge Base Validation deep nets. This approach can be applied to a wide range of information extraction systems as it does not make assumptions about the knowledge representation, language and domain of the data. Experiments clearly show the benefit of using this combined approach on the three different  benchmarks, providing a significant improvement over the state of the art solution based on distantly supervised RE only. The experiments also demonstrate that the proposed system is highly scalable as we were able to apply it to a web scale corpus and hundreds of relations. In addition, we show that the proposed Relation Validation methods are more effective than alternatives based on Probabilistic Soft Logics while they do not require neither ontological constraints nor manually supervised data. 


For the future, we plan to explore the generative aspect of the KBC net such as predicting triples outside the set drawn from IE, with the goal of extracting implicit information from corpora. In addition, we plan to exploit constraints from the ontology within the Knowledge Base Validation and Completion algorithms, to further improve the accuracy of our system and provide reasoner-like behaviors enabling explainability of the decision made by the system. Last, but not least, we plan to explore this methodology to automatically induce KG in the context of enterprise search engines, with the goal of generating infoboxes and a discovery experience over domain specific document collections in any domain. 

All those are just simple steps towards our broader research program around hybridizing symbolic and sub-symbolic methods in AI using deep learning. Our ultimate goal is to enable reading comprehension by acquiring knowledge and inference rules from text and use them to enable reasoning. This work is expected to become a critical part of that bigger mission.

\bibliographystyle{named}
\bibliography{nips_2018}

\end{document}